\documentclass[10pt]{article} 
\usepackage[preprint]{tmlr}

\usepackage{amsmath,amsfonts,bm}

\def\eqref#1{equation~\ref{#1}}

\def\1{\bm{1}}

\DeclareMathAlphabet{\mathsfit}{\encodingdefault}{\sfdefault}{m}{sl}
\SetMathAlphabet{\mathsfit}{bold}{\encodingdefault}{\sfdefault}{bx}{n}

\usepackage{wrapfig}
\usepackage{hyperref}
\usepackage{url}
\usepackage{physics}
\usepackage[utf8]{inputenc} 
\usepackage[T1]{fontenc}    
\usepackage{hyperref}       
\usepackage{url}            
\usepackage{booktabs}       
\usepackage{amsfonts}       
\usepackage{nicefrac}       
\usepackage{microtype}      
\usepackage{xcolor}         
\usepackage{amsmath}
\usepackage[whole]{bxcjkjatype}
\usepackage{graphicx}
\usepackage{color}
\usepackage{bm}
\usepackage{stackengine}
\def\delequal{\mathrel{\ensurestackMath{\stackon[1pt]{=}{\scriptstyle\Delta}}}}
\newcommand{\B}[1]{{\bm #1}} 
\newcommand{\mac}[1]{{\mathcal #1}}
\newcommand{\mab}[1]{{\mathbb #1}}
\definecolor{customred}{RGB}{255,0,29}
\definecolor{customgreen}{RGB}{46,125,57}
\definecolor{customyellow}{RGB}{255,126,42}

\newtheorem{theorem}{Theorem}[section]

\title{Statistical Mechanics of Min-Max Problems}

\author{\name Yuma Ichikawa \email ichikawa-yuma1@g.ecc.u-tokyo.ac.jp, ichikawa.yuma@fujitsu.com \\
      \addr Department of Basic Science, 
      University of Tokyo \\
      Fujitsu Limited
      \AND
      \name Koji Hukushima \email k-hukushima@g.ecc.u-tokyo.ac.jp \\
      \addr Department of Basic Science, University of Tokyo}

\def\month{MM}  
\def\year{YYYY} 
\def\openreview{\url{https://openreview.net/forum?id=XXXX}} 

\begin{document}

\maketitle

\begin{abstract}
    Min-max optimization problems, also known as saddle point problems, have attracted significant attention due to their applications in various fields, such as fair beamforming, generative adversarial networks (GANs), and adversarial learning. 
    However, understanding the properties of these min-max problems has remained a substantial challenge. 
    This study introduces a statistical mechanical formalism for analyzing the equilibrium values of min-max problems in the high-dimensional limit, while appropriately addressing the order of operations for min and max. 
    As a first step, we apply this formalism to bilinear min-max games and simple GANs, deriving the relationship between the amount of training data and generalization error and indicating the optimal ratio of fake to real data for effective learning.
    This formalism provides a groundwork for a deeper theoretical analysis of the equilibrium properties in various machine learning methods based on min-max problems and encourages the development of new algorithms and architectures.
\end{abstract}

\section{Introduction}\label{sec:introduction}
Min-max optimization problems, also known as saddle point problems, are well-known classical optimization problems extensively studied in the context of zero-sum games \citep{wald1945statistical, von2007theory}. 
These problems have diverse applications across various fields, such as game theory, machine learning, and signal processing. 
In game theory, min-max problems arise in zero-sum games where one player's gain corresponds to another's loss. 
Several methods have been proposed to find the min-max value or equilibrium points in these games \citep{dem1972numerical, maistroskii1977gradient, bruck1977weak, lions1978methode, nemirovsky1983wiley, freund1999adaptive}. 
In machine learning, min-max games are relevant for training generative adversarial networks (GANs) \citep{goodfellow2020generative, arjovsky2017wasserstein}, where a generator creates synthetic fake data, and a discriminator distinguishes between fake and real data in a zero-sum game setting. 
Additionally, in adversarial learning, these problems are employed to train models that are robust to adversarial attacks by optimizing a worst-case perturbed loss function \citep{szegedy2013intriguing, goodfellow2014explaining, papernot2016limitations,  madry2017towards}, 

Despite their widespread application of min-max optimization problems, several challenges still need to be addressed, including understanding the usefulness of these min-max formulations, evaluating the convergence properties of the algorithms, and conducting sensitivity analyses of min-max values. 
One promising approach to addressing these issues is to focus on the typical-case behavior of min-max problems in randomized instance ensembles, such as dataset ensembles.
Statistical-mechanical approaches, which have demonstrated their effectiveness in analyzing the typical-case behavior of randomized instance ensembles of optimization and constraint-satisfaction problems \citep{mezard1986replica, fontanari1995statistical}, provide a powerful formalism for such analyses.  
Extending this formalism to analyze the typical-case behavior of min-max values thus presents a potential direction for further research, although this has not yet been fully explored. 
Therefore, this study applies the statistical mechanical formalism to min-max problems, modeling them as a virtual two-temperature system.
This formalism enables sensitivity analysis of the typical-case min-max values in the high-dimensional limit. 
Notably, this formalism properly addresses the order of min-max operations, which is critical in non-convex scenarios where interchanging the order of min and max can lead to incorrect results. 
By using this formalism, we analyze typical-case min-max values of bilinear min-max games and simple GANs.
In particular, we derive the relationship between the amount of training data and generalization error and indicate the optimal ratio of fake data to real data for effective learning.

Our main contributions are as follows:
\begin{itemize}
    \item We introduce a statistical-mechanical formalism developed for sensitivity analysis of equilibrium values in high-dimensional min-max problems.
    \item Applying this approach, we conduct a detailed sensitivity analysis on a bilinear min-max game to verify the theoretical validity of our approach.
    \item Building on this formalism, we analyze the generalization performance of GANs and determine the optimal ratio between fake and real data for effective training.
\end{itemize}

\paragraph{Notation}
Here, we summarize the notations used in this study. 
We use the shorthand expression $[N] = \{1, 2, \ldots, N\}$, where $N \in \mab{N}$. $I_{d} \in \mab{R}^{d \times d}$ denotes a $d \times d$ identity matrix, and $\B{1}_{d}$ denotes the vector $(1, \ldots, 1)^{\top} \in \mab{R}^{d}$ and $\B{0}_{d}$ denotes the vector $(0, \ldots, 0)^{\top} \in \mab{R}^{d}$. 
For the matrix $A = (A_{ij}) \in \mab{R}^{d \times k}$ and a vector $\B{a}=(a_{i}) \in \mab{R}^{d}$, we use the shorthand expressions $dA \delequal 
 \prod_{i=1}^{d} \prod_{j=1}^{k} dA_{ij}$ and $d\B{a} \delequal \prod_{i=1}^{d} da_{i}$, respectively.

\section{Statistical Physics Formalism for Min-Max Optimization Problems}\label{sec:min-max-replica}
The section introduces a statistical-mechanical formalism that models min-max problems as a virtual two-temperature system from a statistical mechanics perspective.
Min-max problems are formally expressed as
\begin{equation}
    \label{eq:min-max-optimization}
    \Omega = \min_{\B{x} \in \mac{X}} \max_{\B{y} \in \mac{Y}} V(\B{x}, \B{y}; A), ~~\mathrm{s.t.}~~\B{x} \in \mac{X} \subseteq \mab{R}^{d_{x}},~~\B{y} \in \mac{Y} \subseteq \mab{R}^{d_{y}}  
\end{equation}
where $V(\cdot, \cdot): \mab{R}^{d_{x}} \times \mab{R}^{d_{y}} \to \mab{R}$ is a bivariate function; $\B{x}\in\mab{R}^{d_{x}}$ and $\B{y} \in \mab{R}^{d_{y}}$ are the optimization variables; $\mac{X}$ and $\mac{Y}$ are the feasible sets; $A$ is a parameter characterizing the problem, e.g., graph $G$.
We introduce the following Boltzmann distribution to analyze min-max problems for a given bivariate function $V(\B{x},\B{y}; A)$ in Eq.~(\ref{eq:min-max-optimization}), with virtual inverse temperatures $\beta_{\min}$ and $\beta_{\max}$:
\begin{equation}
    p_{\beta_{\min}, \beta_{\max}} (\B{x}; A) = \frac{1}{Z(\beta_{\min}, \beta_{\max}, A)} e^{-\beta_{\min} \left(\frac{1}{\beta_{\max}} \ln \int d\B{y} e^{\beta_{\max} V(\B{x}, \B{y}; A)} \right)},
\end{equation}
where $Z(\beta_{\min}, \beta_{\max}, A)$ is the normalization constant, also known as the partition function. Hereafter, we refer to it as the partition function. 
By first taking the limit $\beta_{\max} \to \infty$ followed by $\beta_{\min} \to \infty$, the distribution $\lim_{\beta_{\min} \to +\infty} \lim_{\beta_{\max} \to +\infty} p_{\beta_{\min}, \beta_{\max}} (\B{x}; A)$ concentrates on a uniform distribution over the min-max values. 
Note that the order of these limits is crucial because the min and max operations cannot be interchanged in non-convex and non-concave min-max problems. While a similar formulation has been used in previous work \citep{varga1998minimax}, they simultaneously take the limits of both $\beta_{\min}$ and $\beta_{\max}$ with a fixed ratio $\beta_{\min}/\beta_{\max} = \mathcal{O}((d_{x})^{0}, (d_{y})^{0})$, which does not fully capture the distinct effects of the min and max operations in non-convex settings.
Such an approach generally does not yield accurate results when the function $V(\B{x}, \B{y}; A)$ is non-convex to the variables.

Statistical-mechanical approaches have demonstrated their effectiveness in analyzing the typical-case behavior of randomized instance ensembles of optimization and constraint-satisfaction problems \citep{mezard1986replica, fontanari1995statistical}. 
This work also focuses on evaluating the typical cases of min-max problems characterized by a random parameter $A$. 
Our main objective is to calculate the logarithm of $Z(\beta_{\min}, \beta_{\max}, A)$ averaged over the random variables $A$ in the limit $\beta_{\max} \to \infty$ followed by $\beta_{\min} \to \infty$:
\begin{equation}
    \Omega = \lim_{\beta_{\min} \to \infty} \lim_{\beta_{\max} \to \infty} f(\beta_{\min}, \beta_{\max})
\end{equation}
where
\begin{equation}
    f(\beta_{\min}, \beta_{\max}) = - \frac{1}{\beta_{\min} d_{x}} \mab{E}_{A} \log Z(\beta_{\min}, \beta_{\max}, A),
\end{equation}
referred to as the free energy density.
Setting the ratio of the inverse temperatures as $p = -\nicefrac{\beta_{\min}}{\beta_{\max}}$, this can be rewritten as 
\begin{align}
    \label{eq:non-replica-form}
   f(\beta_{\min}, \beta_{\max}) &= -  \frac{1}{\beta_{\min} d_{x}}\mab{E}_{A} \log \int d\B{x} e^{-\beta_{\min} \left(\frac{1}{\beta_{\max}} \ln \int d\B{y} e^{\beta_{\max} V(\B{x}, \B{y}; A)}\right)}, \notag \\
    &= -  \frac{1}{\beta_{\min} d_{x}}\mab{E}_{A} \log \int d\B{x} \left(\int d\B{y} e^{\beta_{\max} V(\B{x}, \B{y}; A)}\right)^{p}.
\end{align}
Although calculating the expectation value of the logarithm is generally difficult, the replica method provides a resolution:
\begin{equation}
    \label{eq:replica-form}
    \Omega = - \lim_{\beta_{\min} \to \infty} \lim_{\beta_{\max} \to \infty} \lim_{\gamma \to +0} \frac{1}{\beta_{\min} d_{x} \gamma} \log \mab{E}_{A} \left(\int d\B{x} \left(\int d\B{y} e^{\beta_{\max} V(\B{x}, \B{y}); A}  \right)^{p} \right)^{\gamma}.
\end{equation}
Instead of directly handling the logarithmic expression in Eq.~(\ref{eq:non-replica-form}), one calculates the average of the $\gamma$-th and $p$-th power for $\gamma, p \in \mab{N}$, performs an analytic continuation to $\gamma, p \in \mab{R}$ for this expression, and finally takes the limits $\gamma \to + 0$, $\beta_{\min} \to + \infty$ and $\beta_{\max} \to + \infty$. 
Based on this replica ``trick'', the calculation can be simplified to the replicated partition function $Z_\gamma(\beta_{\min},\beta_{\max})$: 
\begin{equation}
\label{eq:analytical-continuos}
    Z_\gamma(\beta_{\min},\beta_{\max}) = \mab{E}_{A} \left(\int d\B{x} \left(\int d\B{y} e^{\beta_{\max} V(\B{x}, \B{y})}  \right)^{p} \right)^{\gamma} = \mab{E}_{A} \int \prod_{a=1}^{\gamma} d\B{x}^{a} \prod_{l=1}^{p} d\B{y}^{al} e^{\beta_{\max} \sum_{a, l} V(\B{x}^{a}, \B{y}^{al}; A)},
\end{equation}
up to the first order of $\gamma$ to take the $\gamma \to +0$ limit in the right hand side of Eq.~(\ref{eq:replica-form}).
This computation is a standard procedure in the statistical physics of random systems and is generally accepted as exact, although rigorous proof has not yet been provided. 

Additionally, before taking the limits, $\beta_{\min}\rightarrow\infty$ and $\beta_{\max}\rightarrow\infty$, the concepts of finite inverse temperature $\beta_{\min}$ and $\beta_{\max}$ correspond to scenarios where neither the minimum nor the maximum is fully achieved, a common situation in the min-max algorithm. 
This approach provides valuable insights into cases where neither extreme is fully realized or both are only partially optimized. Exploring novel algorithms based on this finite-temperature generalization of min-max problems represents an intriguing direction for future work.
Furthermore, in game theory, this formalism can be interpreted as modeling games with relaxed assumptions of complete rationality.
    
In the following sections, we apply this formalism to a fundamental and significant bilinear min-max game, demonstrating that the analytic continuation of $p$ is a rigorous operation. We then analyze the minimal model of GANs as a more practical example.
    
\section{Bilinear Min-max Games}\label{sec:bilinear-min-max-games}
This min-max formalism introduces two replica parameters: $\gamma$, associated with the randomness of $A$, and $p$, related to the dual structure of min-max problems. 
The analytic continuation with respect to the replica parameter $\gamma$ is widely recognized as effective and is frequently employed in the statistical mechanics of optimization. 
However, the analytic continuation of the replica parameter $p$ has not been explored. 
While establishing its mathematical validity presents challenges, this study eliminates the influence of the replica parameter $\gamma$ associated with the randomness of $A$ and rigorously demonstrates that the analytic continuation with respect to $p$ holds for fundamental bilinear min-max games.
Specifically, we show that the free energy density derived using the replica trick in Eq.~(\ref{eq:replica-form}), as explained in Section \ref{sec:min-max-replica}, is equivalent to the exact expression in Eq.~(\ref{eq:non-replica-form}) derived without analytic continuation of $\gamma$ and $p$ for bilinear min-max games \citep{tseng1995linear, daskalakis2017training}.

Bilinear games are regarded as a fundamental example for studying new min-max optimization algorithms and techniques \citep{daskalakis2017training, gidel2019negative, gidel2018variational, liang2019interaction}. 
Mathematically, bilinear zero-sum games can be formulated as the following min-max problem:
\begin{equation}
    \min_{\B{x} \in \{0, 1\}^{d_{x}}} \max_{\B{y} \in \{0, 1\}^{d_{y}}} V(\B{x},\B{y}; \B{W}),
\end{equation}
where $V(\cdot,\cdot)$ is given by
\begin{equation}
    V(\B{x},\B{y}; \B{W}) = \frac{1}{2 d_{x}} \B{x}^{\top} W_{xx} \B{x} + \frac{1}{2 d_{y}} \B{y}^{\top} W_{yy} \B{y} + \frac{1}{\sqrt{d_{x} d_{y}}} \B{x}^{\top} W_{xy} \B{y} + \B{x}^{\top}\B{b}_{x} + \B{y}^{\top}\B{b}_{y},
\end{equation}
where $\B{W}=(W_{xx}, W_{yy}, W_{xy}, \B{b}_{x}, \B{b}_{y})$. 
For simplicity, we assume $W_{xx} = w_{xx} \B{1}_{d_{x} \times d_{x}} \in \mab{R}^{d_{x} \times d_{x}}$, $W_{xy} = w_{xx} \B{1}_{d_{x} \times d_{y}} \in \mab{R}^{d_{x} \times d_{y}}$, $W_{yy} = w_{yy} \B{1}_{d_{y} \times d_{y}} \in \mab{R}^{d_{y} \times d_{y}}$, $\B{b}_{x} = b_{x} \B{1}_{d_{x}} \in \mab{R}^{d_{x}}$, and $\B{b}_{y} = b_{y} \B{1}_{d_{y}} \in \mab{R}^{d_{y}}$. The following results can be readily extended to matrices $W_{xx}$, $W_{yy}$, and $W_{xy}$, which have a limited number of eigenvalues of $\mac{O}(1)$.

In this setting, the analytically continued free energy density $\hat{f}(\beta_{\min}, \beta_{\max}; \B{W})$ calculated using Eq.~(\ref{eq:replica-form}) coincides with the exact free energy density $f(\beta_{\min}, \beta_{\max}; \B{W})$ from Eq.~(\ref{eq:non-replica-form}).
\begin{theorem}
    \label{theorem:exact-case}
    For any $\beta_{\min}, \beta_{\max} \in \mab{R}$ and $w_{xx}, w_{xy}, w_{yy}, b_{x}, b_{y} \in \mab{R}$, the following equality holds:
    \begin{equation*}
        f(\beta_{\min}, \beta_{\max}; \B{W}) = \hat{f}(\beta_{\min}, \beta_{\max}; \B{W})
    \end{equation*}
    where
    \begin{multline*}
        f(\beta_{\min}, \beta_{\max}; \B{W}) = \underset{m^{x}, m^{y}}{\mathrm{extr}} \Bigg[ \frac{w_{xx}}{2} (m^{x})^{2} + \frac{\kappa w_{yy}}{2} (m^{y})^{2} + w_{xy} \kappa^{1/2}  m^{x} m^{y} \\
        + b_{x} m^{x} + \kappa b_{y} m^{y} 
        - \frac{1}{\beta_{\min}} H(m^{x})+ \frac{\kappa}{\beta_{\max}}H(m^{y})\Bigg] 
    \end{multline*}
    where $\kappa=\nicefrac{d_{y}}{d_{x}}$, $H(x)=-x \log(x) -(1-x) \log (1-x)$ denotes binary cross entropy, and $\mathrm{extr}$ denotes the extremum operation.
\end{theorem}
This theorem establishes the validity of the analytic continuation for the replica parameter $p$ using Eq.~(\ref{eq:analytical-continuos}) for bilinear min-max games. 
The detailed proof of this theorem is provided in Appendix.~\ref{sec:proof-theorem}. 

\section{Generative Adversarial Networks}\label{sec:gan}
Generative adversarial networks (GANs) \citep{goodfellow2020generative} aim to model high-dimensional probability distributions based on training datasets. 
Despite significant progress in practical applications \citep{arjovsky2017wasserstein, lucic2018gans, ledig2017photo, isola2017image, reed2016generative}, several issues are yet to be resolved, including how the amount of training data influences generalization performance and how sensitive GANs are to specific hyperparameters. 
This section analyzes the relationship between the amount of training data and generalization error.
Additionally, we conduct a sensitivity analysis on the ratio of fake data generated by the generator to the amount of training data, which is critical for the training of GANs.
Our analysis employs a minimal setup that captures the intrinsic structure and learning dynamics of GANs \citep{wang2019solvable}.
We consider the high-dimensional limit, where the number of real and fake samples, $n$ and $\tilde{n}$, and the dimension $d$ are large while remaining comparable. 
Specifically, we analyze the regime in which $n, \tilde{n}, d \to \infty$ while maintaining a comparable ratio, i.e., $\alpha = \nicefrac{n}{d} = \Theta(d^{0})$ and $\tilde{\alpha} = \nicefrac{\tilde{n}}{d} = \Theta(d^{0})$, commonly referred to as sample complexity.

\subsection{Settings}\label{subsec:setting-gan}
\paragraph{Generative model for the dataset}
We consider that a training dataset $\mac{D} = \{\B{x}^{\mu} \}_{\mu=1}^{n}$, where each $\B{x}^{\mu} \in \mab{R}^{d}$ is drawn by the following distribution:
\begin{equation}
    \label{eq:gan-dataset-dist}
    \B{x}^{\mu} = \frac{1}{\sqrt{d}} \B{w}^{\ast} c^{\mu} + \sqrt{\eta} \B{n}^{\mu},
\end{equation}
where $\B{w}^{\ast} \in \mab{R}^{d}$ is a deterministic feature vector, $c^{\mu} \in \mab{R}$ is random scalar drawn from a standard normal distribution $p(c) = \mac{N}(0, 1)$, $\B{n}^{\mu}$ is a background noise vector whose components are i.i.d from the standard normal distribution $\mac{N}(\B{0}_{d}, I_{d})$, and $\eta \in \mab{R}$ is a scalar parameter to control the strength of the noise. 
We also assume that $\|\B{w}^{\ast}\|^{2}=1$.
This generative model, known as the spiked covariance model \citep{wishart1928generalised, potters2020first}, has been studied in statistics to analyze the performance of unsupervised learning methods such as PCA \citep{ipsen2019phase, biehl1993statistical, hoyle2004principal}, sparse PCA \citep{lesieur2015phase}, deterministic autoencoders \citep{refinetti2022dynamics}, and variational autoencoder \citep{pmlr-v238-ichikawa24a, ichikawa2023dataset}. 

\paragraph{GAN model}
Following \citet{wang2019solvable}, 
we assume that the generator has the same linear structure as the dataset generative model described in Eq.~(\ref{eq:gan-dataset-dist}):
\begin{equation}
    \label{eq:generator-data}
    g(z; \B{w}) = \frac{1}{\sqrt{d}} \B{w} z + \sqrt{\tilde{\eta}} \tilde{\B{n}},
\end{equation}
where $\B{w} \in \mab{R}^{d}$ is a learnable parameter, $z \in \mab{R}$ is latent variable drawn from a standard normal distribution $p(z) = \mac{N}(0, 1)$, $\tilde{\B{n}}$ is a noise vector whose components are i.i.d from the standard normal distribution $\mac{N}(\B{0}_{d}, I_{d})$, and $\tilde{\eta} \in \mab{R}$ is a scalar parameter to control the strength of the noise.

We also define the linear discriminator as 
\begin{equation}
    \label{eq:discriminator}
    d(\B{x}; \B{v}) = f \left( \frac{1}{\sqrt{d}} \B{v}^{\top} \B{x} \right), 
\end{equation}
where $\B{x}$ is an input vector, which can be either the real data $\B{x}^{\mu}$ from Eq.~(\ref{eq:gan-dataset-dist}) or the fake one $g(z^{\tilde{\mu}}; \B{w})$ from Eq.~(\ref{eq:generator-data}).
The vector $\B{v} \in \mab{R}^{d}$ is a learnable parameter, and $f: \mab{R} \to \mab{R}$ can be any function.

\paragraph{Training algorithm}
The GAN is trained by solving the following min-max optimization problem:
\begin{equation}
    \label{eq:gan-min-max-problem}
    \min_{\B{w} \in \mab{R}^{d}} \max_{\B{v} \in \mab{R}^{d}} V(\B{w}, \B{v}; \mac{D}),
\end{equation}
where
\begin{align}
    \label{eq:gan-value-func}
    V(\B{w}, \B{v}; \mac{D}, \tilde{\mac{D}}) &= \sum_{\mu=1}^{n} \phi\left(d(\B{x}^{\mu}; \B{v})\right) - \sum_{\tilde{\mu}=1}^{\tilde{n}} \tilde{\phi}\left(d(g(z^{\tilde{\mu}}; \B{w}); \B{v}) \right) - \frac{\lambda}{2} \|\B{v}\|^{2} + \frac{\tilde{\lambda}}{2} \|\B{w}\|^{2}, 
\end{align}
where $\tilde{\mac{D}}=\{z^{\mu}\}_{\tilde{\mu}=1}^{\tilde{n}}$ is the latent values of the fake data. The last two terms are regularization terms, where $\lambda$ and $\tilde{\lambda}$ control the regularization strength.
As we assumes a linear discriminator, $V(\B{w}, \B{v}; \mac{D}, \tilde{\mac{D}})$ can be expressed as
\begin{align}
    V(\B{w}, \B{v}; \mac{D}, \tilde{\mac{D}}) &= \sum_{\mu=1}^{n} \phi\left(\frac{1}{\sqrt{d}} \B{v}^{\top} \B{x}^{\mu} \right) - \sum_{\tilde{\mu}=1}^{\tilde{n}} \tilde{\phi}\left(\frac{1}{\sqrt{d}} \B{v}^{\top} g(z^{\tilde{\mu}}; \B{w}) \right) - \frac{\lambda}{2} \|\B{v}\|^{2} + \frac{\tilde{\lambda}}{2} \|\B{w}\|^{2},
\end{align}
This value function defined in Eq.~(\ref{eq:gan-value-func}) is a general form that includes various types of GANs. Specifically, when $\phi = \tilde{\phi}$ and $ \|\phi\|_{L} \leq 1 $, it represents a Wasserstein GANs (WGANs) \citep{arjovsky2017wasserstein} and, when $\phi(x) = \log \sigma(x) $ and $\tilde{\phi}(x) = - \log (1-\sigma(x))$ with $\sigma $ being the sigmoid function, it corresponds to the Vanilla GANs, which minimize the JS-divergence \citep{goodfellow2014generative}.

\paragraph{Generalization error}
In the ideal case where the generator perfectly learns the underlying true probability distribution, we have $\B{w}^{\ast}=\B{w}$.
Therefore, we define the generalization error $\varepsilon_{g}$ as 
\begin{equation}
    \label{eq:main-generalization-error}
    \varepsilon_{g}(\bar{\B{w}}, \B{w}^{\ast}) = \frac{1}{d} \mab{E}_{\mac{D}}\left[ \|\bar{\B{w}} - \B{w}^{\ast}\|^{2} \right],
\end{equation}
where $\bar{\B{w}}$ denotes the min-max optimal value in Eq.~(\ref{eq:gan-min-max-problem}). 
The generalization error, $\varepsilon_{g}$, quantifies the accuracy of signal recovery from the training data.

\subsection{Replica Calculation}
We apply the replica formalism sketched in Section \ref{sec:min-max-replica} to derive a set of deterministic equations characterizing the typical behavior of GANs.

In this problem setting, the replicated partition function $Z_\gamma$ in Eq. (\ref{eq:analytical-continuos}) can be expressed as
\begin{align*}
    Z_\gamma            
    &= \int  \prod_{a=1}^{\gamma} d\B{w}^{a} \prod_{l=1}^{p} d\B{v}^{al} \left(\mab{E}_{c, \B{n}} e^{\beta_{\max} \sum_{al} \phi\left(\frac{1}{d} (\B{v}^{al})^{\top} \B{w}^{\ast} c + \sqrt{\frac{\eta}{d}} (\B{v}^{al})^{\top} \B{n} \right)} \right)^{n} \\
    &\times \left(\mab{E}_{z} e^{-\beta_{\max} \sum_{al}  \tilde{\phi}\left(\frac{1}{d} (\B{v}^{al})^{\top} \B{w}^{a} z + \sqrt{\frac{\tilde{\eta}}{d}} (\B{v}^{al})^{\top} \tilde{\B{n}}  \right)} \right)^{\tilde{n}}  e^{\frac{\beta_{\max}}{2} \sum_{al} \left(\tilde{\lambda} \|\B{w}^{a}\|^{2} - \lambda \|\B{v}^{al}\|^{2} \right)}. 
\end{align*}
To take the average over $\B{n}$ and $\tilde{\B{n}}$, we notice that since $\B{n}$ and $\tilde{\B{n}}$ follow a multivariate normal distribution $\mac{N}(\B{0}_{d}, I_{d})$, the quantities $\B{u} = ((\B{v}^{al})^{\top} \B{n}/\sqrt{d})_{a, l}$ and $\tilde{\B{u}} = ((\B{v}^{al})^{\top} \tilde{\B{n}}/\sqrt{d})_{a, l}$ also follow a Gaussian multivariate distribution as
\begin{equation*}
    p(\B{u}) = p(\tilde{\B{u}}) = \mac{N}(\B{0}_{\gamma p}, \B{Q}),
\end{equation*}
where
\begin{equation*}
    \B{Q} = (Q_{ls}^{ab}) \in \mab{R}^{\gamma p \times \gamma p},~~Q_{ls}^{ab} = \frac{1}{d} (\B{v}^{al})^{\top} \B{v}^{bs}.
\end{equation*}
To conduct further computations, we introduce auxiliary variables through the following identities: 
\begin{align*}
    &1 = \prod_{abls} d \int \delta (dQ_{ls}^{ab} - (\B{v}^{al})^{\top} \B{v}^{bs}) dQ_{ls}^{ab} = \prod_{al} d \int \delta (d m^{a}_{l} - (\B{v}^{al})^{\top} \B{w}^{\ast}) dm^{a}_{l} = \prod_{al} d \int \delta (d b^{a}_{l} - (\B{v}^{al})^{\top} \B{w}^{a}) db^{a}_{l}. 
\end{align*}
The replicated partition function can then be expressed as 
\begin{align*}
    Z_\gamma & = \int d\B{Q} d\B{m} d\B{b} e^{\beta_{\min} d (\mac{S}(\B{Q}, \B{m}, \B{b}) + \mac{T}(\B{Q}, \B{m}, \B{b}))}, 
\end{align*}
where we define the entropic term $\mac{S}(\B{Q}, \B{m}, \B{b})$ and energetic term $\mac{T}(\B{Q}, \B{m}, \B{b})$ as follows:
\begin{align*}
    \mac{S}(\B{Q}, \B{m}, \B{b}) 
    & \delequal \frac{1}{d \beta_{\min}} \ln \int \prod_{al} d\B{w}^{a} d\B{v}^{al} \prod_{abls} d \int \delta (dQ_{ls}^{ab} - (\B{v}^{al})^{\top} \B{v}^{bs}) \\
    &\times \prod_{al} d \int \delta (d m^{a}_{l} - (\B{v}^{al})^{\top} \B{w}^{\ast}) d \int \delta (d b^{a}_{l} - (\B{v}^{al})^{\top} \B{w}^{a}) e^{\frac{\beta_{\max}}{2}\sum_{al}(\tilde{\lambda} \|\B{w}^{a}\|^{2}-\lambda \|\B{v}^{al}\|^{2})}, \\
    \mac{T}(\B{Q}, \B{m}, \B{b}) 
    & \delequal \frac{\alpha}{\beta_{\min}}  \ln \left(\int Dc \int d\B{u} p(\B{u}) e^{\beta_{\max} \sum_{al} \phi\left(\frac{1}{d} (\B{v}^{al})^{\top} \B{w}^{\ast} c + \sqrt{\frac{\eta}{d}} (\B{v}^{al})^{\top} \B{n} \right)} \right) \\
    &+ \frac{\tilde{\alpha}}{\beta_{\min}} \ln \left(\int Dz \int d\tilde{\B{u}} p(\tilde{\B{u}}) e^{-\beta_{\max} \sum_{al}  \tilde{\phi}\left(\frac{1}{d} (\B{v}^{al})^{\top} \B{w}^{a} z + \sqrt{\frac{\tilde{\eta}}{d}} (\B{v}^{al})^{\top} \tilde{\B{n}}  \right)} \right).
\end{align*}
Using the Fourier representation of the delta function, $\mac{S}(\B{Q}, \B{m}, \B{b})$ is further expressed as 
\begin{multline}
    \label{eq:gan-entropy-1}
    \mac{S}(\B{Q}, \B{m}, \B{b})=\frac{1}{d \beta_{\min}}\log \int d\tilde{\B{Q}}d\tilde{\B{m}} d\tilde{\B{b}} e^{d \left(\frac{1}{2} \mathrm{tr} \tilde{\B{Q}} \B{Q} - \tilde{\B{m}}^{\top} \B{m} - \tilde{\B{b}}^{\top} \B{b} \right)} \\
    \left(\int \prod_{al} dw^{a} dv^{al} e^{-\frac{1}{2} \sum_{abls} \tilde{Q}^{ab}_{ls} v^{al} v^{bs} + w^{\ast} \sum_{al} \tilde{m}_{l}^{a} v^{al} + \sum_{al} \tilde{b}^{a}_{l} w^{a} v^{al} + \frac{\beta_{\max}}{2} \sum_{al} \left( \tilde{\lambda} (w^{a})^{2} - \lambda (v^{al})^{2}  \right)}\right)^{d}.
\end{multline}

\paragraph{Replica symmetric ansatz} 
Here, we assume the following symmetric structure:
\begin{align}
&\forall a, b \in [\gamma], \forall l, s \in [p], ~Q_{ls}^{ab} = q + \frac{\Delta}{\beta_{\min}} \delta_{ab}  + \frac{\chi}{\beta_{\max}} \delta_{ls}  \delta_{ab},  \label{eq:rs-q}\\
&\forall a, b \in [\gamma], \forall l, s \in [p], ~\tilde{Q}_{ls}^{ab} = \beta_{\max} \hat{q} \delta_{ls} \delta_{ab} -\frac{\beta_{\max}^{2}}{\beta_{\min}}  \hat{\Delta} \delta_{ab} - \beta_{\max}^{2} \hat{\chi}, \label{eq:rs-con-q}\\
&\forall a\in [\gamma], l \in [p],~~m_{l}^{a} = m,~~\tilde{m}_{l}^{a} = \beta_{\max} \hat{m}, \label{eq:rs-m} \\
&\forall a\in [\gamma], l \in [p],~~b_{l}^{a} = b,~~\hat{b}_{l}^{a}=\beta_{\max} \hat{b}. \label{eq:rs-b}
\end{align}
This replica symmetric (RS) structure restricts the integration of the replicated weight parameters $\{\B{w}^{a}\}$, $\{\B{v}^{al}\}$ across the entire $\mab{R}^{(d \times \gamma p)} \times \mab{R}^{(d \times \gamma p)}$ to a subspace that satisfies the constraints in Eq.~(\ref{eq:rs-q})-(\ref{eq:rs-b}). 
This structure, along with scaling by the maximum and minimum beta values, is similar to the standard one-step replica symmetry breaking (1RSB) \citep{mezard1987spin, takahashi2022macroscopic}. 

We now turn to the entropic term $\mac{S}(\B{Q}, \B{m}, \B{b})$. 
The terms that exclude the integrals with respect to $\{\B{v}^{al}\}$ and $\{\B{w}^{a}\}$ can be expressed as
\begin{align}
    &\frac{1}{2} \mathrm{tr} \tilde{\B{Q}} \B{Q} - \tilde{\B{m}}^{\top} \B{m} - \tilde{\B{b}}^{\top} \B{b} \nonumber \\
    &= \gamma \beta_{\min} \left(- \frac{1}{2} \left(\hat{q} \left(q+\frac{\Delta}{\beta_{\min}} + \frac{\chi}{\beta_{\max}}\right) - \chi \left(\hat{\chi} + \frac{\hat{\Delta}}{\beta_{\min}} \right) + \hat{\chi} \Delta + \hat{\Delta}q + \frac{\Delta \hat{\Delta}}{\beta_{\min}} \right) + \hat{m} m + \hat{b} b \right).  
\end{align}
The term that includes the integrals with respect to $\{\B{v}^{al}\}$ and $\{\B{w}^{a}\}$ can be expressed as
\begin{align*}
    &\mab{E}_{z} \int \prod_{al} dw^{a} D\zeta^{a} dv^{al} e^{-\frac{1}{2} \beta_{\max} (\hat{q}+\lambda) \sum_{al}(v^{al})^{2} + \beta_{\max} \sum_{al} \left(\sqrt{\frac{\hat{\Delta}}{\beta_{\min}}} \zeta^{a} + \sqrt{\hat{\chi}} z + w^{\ast} \hat{m} + w^{a} \hat{b} \right) v^{al} - \frac{\tilde{\lambda}  \beta_{\min}}{2} \sum_{a} (w^{a})^{2}}, \\
    &=\mab{E}_{z} \int \prod_{a} dw^{a} D\zeta^{a}  \left(\int dv^{a} e^{-\frac{1}{2} \beta_{\max} (\hat{q}+\lambda) (v^{a})^{2} + \beta_{\max}  \left(\sqrt{ \frac{\hat{\Delta}}{\beta_{\min}}} \zeta^{a} + \sqrt{\hat{\chi}} z + w^{\ast} \hat{m} + w^{a} \hat{b} \right) v^{a} }\right)^{p} e^{- \frac{\tilde{\lambda} \beta_{\min}}{2} \sum_{a} (w^{a})^{2}}, \\
    &=\mab{E}_{z} \int \prod_{a} dw^{a} D\zeta^{a} e^{-\frac{\tilde{\lambda} \beta_{\min}}{2} \sum_{a} (w^{a})^{2} - \frac{\beta_{\min}}{2 (\hat{q}+ \lambda)}\sum_{a} \left(\sqrt{\frac{\hat{\Delta}}{\beta_{\min}}} \zeta^{a} + \sqrt{\hat{\chi}} z + w^{\ast} \hat{m} + w^{a} \hat{b}  \right)^{2} }, \\
    &= \mab{E}_{z} \left(\int dw d\zeta e^{\beta_{\min}\left(-\frac{1}{2} \zeta^{2} - \frac{\tilde{\lambda}}{2} w^{2} - \frac{(\sqrt{\hat{\Delta}} \zeta + \sqrt{\hat{\chi}} z + w^{\ast} \hat{m} + w \hat{b})^{2}}{2(\hat{q} + \lambda)}   \right)}  \right)^{\gamma}. 
\end{align*}
This can be derived using the identity, for any $a \in \mab{R}_{+}$ and any $x \in \mab{R}$, $e^{\frac{a}{2} x^{2}} = \int Dz e^{\sqrt{a} z x}$.
Summarizing these results, the entropic term can be written as 
\begin{multline*}
    \mac{S}(\B{Q}, \B{m}, \B{b}, \tilde{\B{Q}}, \tilde{\B{m}}, \tilde{\B{b}}) = \gamma  \left(- \frac{1}{2} \left(\hat{q} \left(q+\frac{\Delta}{\beta_{\min}} + \frac{\chi}{\beta_{\max}}\right) - \chi \left(\hat{\chi} + \frac{\hat{\Delta}}{\beta_{\min}} \right) + \hat{\chi} \Delta + \hat{\Delta}q + \frac{\Delta \hat{\Delta}}{\beta_{\min}}  \right) + \hat{m} m + \hat{b} b \right)  \\
    + \frac{1}{\beta_{\min}} \int Dz \log \int dw d\zeta e^{\beta_{\min}\left(-\frac{1}{2} \zeta^{2} - \frac{\tilde{\lambda}}{2} w^{2} - \frac{(\sqrt{\hat{\Delta}} \zeta + \sqrt{\hat{\chi}} z + w^{\ast} \hat{m} + w \hat{b})^{2}}{2(\hat{q} + \lambda)} \right)} \Bigg).
\end{multline*}
By taking the limit as $\beta_{\max} \to \infty$ followed by $\beta_{\min} \to \infty$, owe obtain
\begin{equation}
    \mac{S}(\B{Q}, \B{m}, \B{b}, \tilde{\B{Q}}, \tilde{\B{m}}, \tilde{\B{b}}) = - \frac{\gamma}{2} \Bigg( q(\hat{q}+\hat{\Delta})-(\chi-\Delta) \hat{\chi} - 2 m \hat{m} - 2 b\hat{b} + \frac{\tilde{\lambda} (\hat{m}^{2}+ \hat{\chi})}{\hat{b}^{2} + (\hat{q}+\hat{\Delta} + \lambda) \tilde{\lambda}}  \Bigg). 
\end{equation}

We next turn to the energetic term $\mac{T}(\B{Q}, \B{m}, \B{b})$.
Under the RS ansatz, $\B{u}$ follows
\begin{align*}
    u^{al} = \sqrt{\frac{\chi}{\beta_{\max}}} x^{al} + \sqrt{\frac{\Delta}{\beta_{\min}}} y^{a} + \sqrt{q} \xi, 
\end{align*}
where $\tilde{x}^{al}$, $x^{al}$, $\tilde{y}^{al}$, $y^{al}$, $\xi$, and $\tilde{\xi}$ follow the standard normal distribution $\mac{N}(0, 1)$. Then, the energetic term $\mac{T}(\B{Q}, \B{m}, \B{b})$ can be expand as
\begin{multline*}
    \mac{T}(\B{Q}, \B{m}, \B{b}) \delequal \underbrace{\frac{\alpha}{\beta_{\min}}  \ln \left(\int Dc \int d\B{u} p(\B{u}) e^{\beta_{\max} \sum_{al} \phi\left(\frac{1}{d} (\B{v}^{al})^{\top} \B{w}^{\ast} c + \sqrt{\frac{\eta}{d}} (\B{v}^{al})^{\top} \B{n} \right)} \right)}_{(a)} \\
    + \underbrace{\frac{\tilde{\alpha}}{\beta_{\min}} \ln \left(\int Dz \int d\tilde{\B{u}} p(\tilde{\B{u}}) e^{-\beta_{\max} \sum_{al}  \tilde{\phi}\left(\frac{1}{d} (\B{v}^{al})^{\top} \B{w}^{a} z + \sqrt{\frac{\tilde{\eta}}{d}} (\B{v}^{al})^{\top} \tilde{\B{n}}  \right)} \right)}_{(b)}. 
\end{multline*}
The term (a) can be simplified as 
\begin{align*}
    \text{(a)} 
    &=\frac{\alpha}{\beta_{\min}} \ln \mab{E}_{c, \xi} \int \prod_{al} Dy^{a} Dx^{al} e^{\beta_{\max} \sum_{al} \phi\left(mc + \sqrt{\eta} \left(\sqrt{\frac{\chi}{\beta_{\max}}} x^{al} + \sqrt{\frac{\Delta}{\beta_{\min}}} y^{a} + \sqrt{q} \xi \right)\right)}, \\
    &=\frac{\alpha}{\beta_{\min}} \ln \mab{E}_{c, \xi} \left(\int Dy \left(\int Dx e^{\beta_{\max} \phi\left(mc + \sqrt{\eta} \left(\sqrt{\frac{\chi}{\beta_{\max}}} x + \sqrt{\frac{\Delta}{\beta_{\min}}} y + \sqrt{q} \xi \right)\right)}  \right)^{p}\right)^{\gamma}, \\
    &=\frac{\alpha}{\beta_{\min}} \gamma \mab{E}_{c, \xi} \log \int Dy \left(\int Dx e^{\beta_{\max} \phi\left(mc + \sqrt{\eta} \left(\sqrt{\frac{\chi}{\beta_{\max}}} x + \sqrt{\frac{\Delta}{\beta_{\min}}} y + \sqrt{q} \xi \right)\right)} \right)^{p} + \mac{O}(\gamma^{2}), \\
    &=\frac{\alpha}{\beta_{\min}} \gamma \mab{E}_{c, \xi} \log \int dy e^{-\frac{\beta_{\min}}{2} y^{2}} \left(\int dx e^{-\frac{\beta_{\max}}{2} x^{2}+ \beta_{\max} \phi\left(mc + \sqrt{\eta} \left(\sqrt{\chi} x + \sqrt{\Delta} y + \sqrt{q} \xi \right)\right)} \right)^{p} + \mac{O}(\gamma^{2}).
\end{align*}
Taking the limit as $\beta_{\max} \to \infty$ followed by $\beta_{\min} \to \infty$, we obtain: 
\begin{equation*}
    (a) = \alpha \gamma  \mab{E}_{c, \xi} \max_{y} \left[-\frac{1}{2} y^{2} - \max_{x} \left[-\frac{1}{2} x^{2} + \phi\left(mc + \sqrt{\eta} \left(\sqrt{\chi} x + \sqrt{\Delta} y + \sqrt{q} \xi \right)\right) \right]  
 \right].
\end{equation*}
Similarly, the term (b) is also expressed as 
\begin{align*}
    (b) &= \frac{\tilde{\alpha}}{\beta_{\min} \gamma}  \mab{E}_{z, \tilde{\xi}} \ln \int d\tilde{y} e^{-\frac{\beta_{\min}}{2} y^{2}} \left(\int d\tilde{x} e^{-\frac{\beta_{\max}}{2} \tilde{x}^{2}- \beta_{\max} \tilde{\phi}\left(bz + \sqrt{\eta} \left(\sqrt{\chi} \tilde{x} + \sqrt{\Delta} \tilde{y} + \sqrt{q} \tilde{\xi} \right)\right)} \right)^{p} + \mac{O}(\gamma^{2}).
\end{align*}
Taking the same limits, we find: 
\begin{equation*}
    (b)= \tilde{\alpha} \gamma \mab{E}_{z, \tilde{\xi}} \max_{\tilde{y}} \left[-\frac{1}{2} y^{2} - \max_{\tilde{x}} \left[-\frac{1}{2} \tilde{x}^{2} - \tilde{\phi}\left(bz + \sqrt{\tilde{\eta}} \left(\sqrt{\chi} \tilde{x} + \sqrt{\Delta} \tilde{y} + \sqrt{q} \tilde{\xi} \right)\right) \right] \right].
\end{equation*}
Putting the entropic term and energetic term together, the free energy density is given by
\begin{equation}
    f = \underset{\substack{\hat{q}, \hat{\chi}, \hat{m}, \hat{b} \\ q, \delta, \chi, m, b}}{\mathrm{extr}} \frac{1}{2} \Bigg( q \hat{q} - (\chi-\Delta) \hat{\chi} - 2 (m \hat{m} + b\hat{b}) + \frac{\tilde{\lambda} (\hat{m}^{2} + \hat{\chi})}{\hat{b}^{2} + (\hat{q}+ \lambda) \tilde{\lambda}}
    - 2 (\alpha \Phi(q, \Delta, \chi, m, b) + \tilde{\alpha} \tilde{\Phi}(q, \Delta, \chi, m, b))  \Bigg),
\end{equation}
where 
\begin{align}
    &\Phi(q, \Delta, \chi, m, b) = \mab{E}_{c, \xi} \max_{y} \left[-\frac{1}{2} y^{2} - \max_{x} \left[-\frac{1}{2} x^{2} + \phi\left(mc + \sqrt{\eta} \left(\sqrt{\chi} x + \sqrt{\Delta} y + \sqrt{q} \xi \right)\right) \right] \right], 
    \label{eq:true-energy-term}\\
    &\tilde{\Phi}(q, \Delta, \chi, m, b) = \mab{E}_{z, \tilde{\xi}} \max_{\tilde{y}} \left[-\frac{1}{2} \tilde{y}^{2} - \max_{\tilde{x}} \left[-\frac{1}{2} \tilde{x}^{2} - \tilde{\phi}\left(bz+ \sqrt{\tilde{\eta}} \left(\sqrt{\chi} \tilde{x} + \sqrt{\Delta} \tilde{y} + \sqrt{q} \tilde{\xi} \right)\right) \right] 
    \right]. 
    \label{eq:false-energy-term}
\end{align}
Note that the min and max operations are involved in the two-level optimization described in Eqs.~(\ref{eq:true-energy-term}) and (\ref{eq:false-energy-term}).

\subsection{Results: Application to Simple GANs}\label{sec:results}
In this subsection, following \citet{wang2019solvable}, we apply the formulation derived above to the simple WGAN to demonstrate its generalization properties and conduct a sensitivity analysis of the ratio $r$ fake to real data.

\paragraph{Self-consistent Equations}
We consider the case where the functions $\phi(x)$ and $\tilde{\phi}(x)$ are both quadratic, defined as $\phi(x) = \tilde{\phi}(x) = x^{2}/2$. This setting allows for an explicit calculation of the free energy density, which is given by
\begin{equation}
    \label{eq:free-energy-simple-WGAN}
    f = \underset{\substack{q, \chi, m, b \\ \hat{q}, \hat{\chi}, \hat{m}, \hat{b}}}{\mathrm{extr}} \frac{1}{2} \Bigg( q \hat{q} - \chi \hat{\chi} - 2 m \hat{m} - 2 b\hat{b} + \frac{\tilde{\lambda} (\hat{m}^{2} + \hat{\chi})}{\hat{b}^{2} + (\hat{q}+ \lambda) \tilde{\lambda}} 
    - \frac{\alpha (\eta q + m^{2})}{\eta \chi -1} - \frac{\tilde{\alpha}(\tilde{\eta} q + b^{2})}{\tilde{\eta} \chi +1}  \Bigg). 
\end{equation}
To find the extremum in Eq.~(\ref{eq:free-energy-simple-WGAN}), we require that the gradient with respect to each order parameter equals zero.
This results in the following set of self-consistent equations:
\begin{multline*}
    q = \frac{\tilde{\lambda}^{2}(\hat{m}^{2}+ \hat{\chi})}{(\hat{b}^{2}+\tilde{\lambda} (\hat{q}+\lambda))^{2}},~~\chi=\frac{\tilde{\lambda}}{\hat{b}^{2}+\tilde{\lambda} (\hat{q}+\lambda)},~~m = \frac{\hat{m} \tilde{\lambda}}{\hat{b}^{2}+\tilde{\lambda} (\hat{q}+\lambda)},~~b = -\frac{\hat{b}\tilde{\lambda}(\hat{m}^{2}+\hat{\chi})}{(\hat{b}^{2} + \tilde{\lambda} (\hat{q}+\lambda))^{2}}, \\
    \hat{q}= \frac{\alpha \eta }{\eta \chi-1}+\frac{\tilde{\alpha}\tilde{\eta}}{\tilde{\eta} \chi+1},~~\hat{\chi} = \frac{\alpha \eta (q \eta + m^{2} \rho)}{(\eta \chi-1)^{2}}+
    \frac{\tilde{\alpha} \tilde{\eta} (q \tilde{\eta} + d^{2}\rho)}{(\tilde{\eta} \chi+1)^{2}},~~\hat{m} =  \frac{\alpha m}{\eta \chi-1},~~\hat{b} = -  \frac{ \tilde{\alpha}d}{\tilde{\eta} \chi+1}.
\end{multline*}
\begin{figure}
    \centering
    \includegraphics[width=0.9\linewidth]{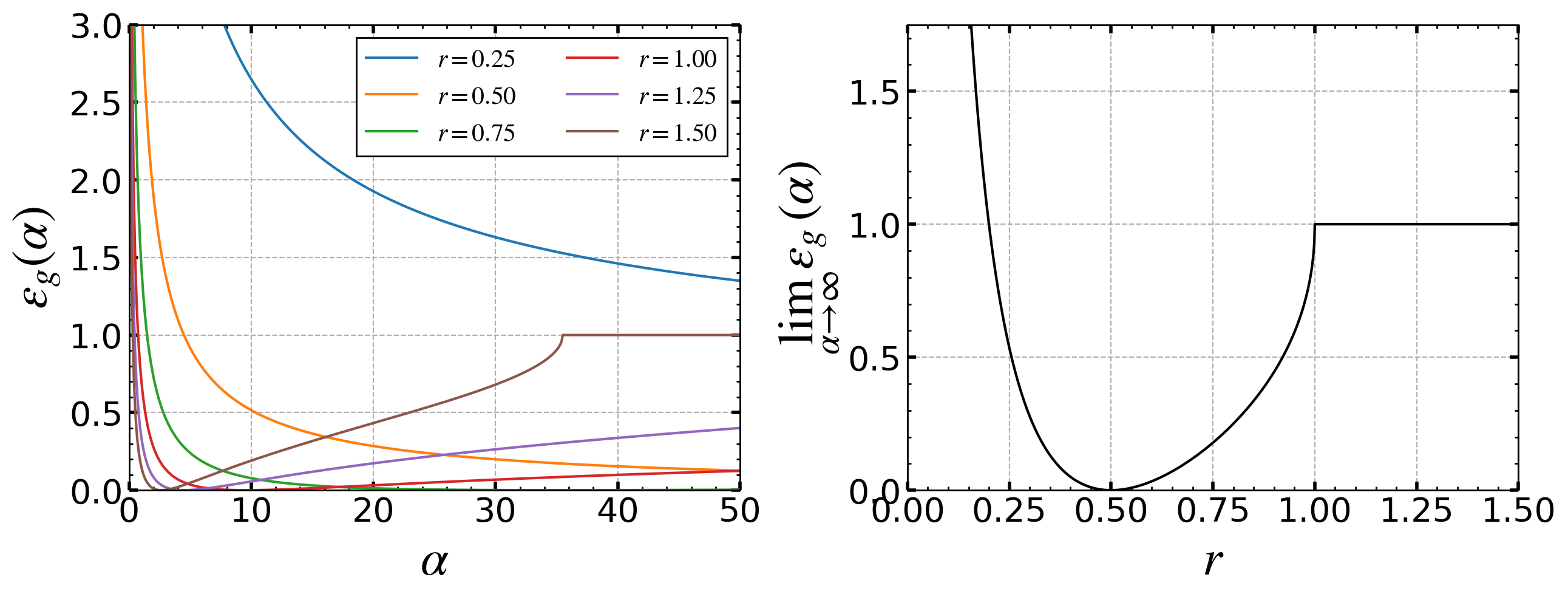}
    \caption{(Left) Generalizatioin error as a function of sample complexity $\alpha$ for different values of the ratio $r$. (Right) Asymptotic generalization error $\lim_{\alpha \to \infty} \varepsilon(\alpha)$ as a function of the the ratio $r$.}
    \label{fig:w-gan-error}
\end{figure}

\paragraph{Learning Curve}
For simplicity, we set $\tilde{\alpha}=r\alpha$ and $\lambda= \tilde{\lambda} = \eta= \tilde{\eta} = 1$. 
Our analysis focuses on how the generalization error depends on $\alpha$ while varying the ratio $r$, as generating fake data from the generator is generally much easier than collecting real data.
Fig.~\ref{fig:w-gan-error} (Left) shows the dependence of generalization error on sample complexity $\alpha$ for various values of the ratio $r$. 
The results demonstrate a sharp decline in the generalization error as the ratio $r$ increases.
However, when $r$ becomes large, the generalization error increases in the region where $\alpha$ is large, eventually leading to a phase where no learning occurs, and the generalization error equals $1$.
This implies that as $\alpha$ increases, the learning becomes dominated by only fake data. 

In contrast, for smaller $r$, real data consistently dominates the objective function $V(\B{w}, \B{v}; \mac{D}, \tilde{\mac{D}})$, resulting in a steady decrease in generalization error. However, the reduced influence of the fake data component in the objective function, which drives the learning of the generator, requires a significantly larger amount of real data for effective generator training.

\paragraph{Asymptotic Generalisation Error}
We next analyze the asymptotic behavior of the generalization error when the sample complexity $\alpha$ becomes sufficiently large. 
The asymptotic behavior of the generalization error as a function of $\alpha$ is given by
\begin{equation*}
    \varepsilon_{g} = 
    \begin{cases}
    \frac{1-2 \sqrt{\frac{1-r}{r}} r}{r} + \frac{2 \sqrt{2} \left(\sqrt{\frac{1-r}{r}} r+r-1\right)}{(r-1) r \alpha^{1/2}} + \mac{O}(\alpha^{-1}) & r \le 1, \\
    1 + \mac{O}(\alpha^{-1}) & r > 1. 
    \end{cases}
\end{equation*}
The results for $\alpha \to \infty$ are shown in Fig.~\ref{fig:w-gan-error}(Right).
The optimal ratio is $r = \nicefrac{1}{2}$, indicating that using fake data approximately equal to half of the real data is effective when the dataset approaches infinity.
At $r=1$, a phase transition occurs, suggesting that the model changes from a phase of effective learning phase to one where fake data becomes dominant. 
Beyond this point, for $r \geq 1$, the model fails to learn any meaningful signal $\B{w}^{\ast}$, and the generalization error is $1$. 

Furthermore, when $r = \nicefrac{1}{2}$, the generalization error scales as $\varepsilon_{g} \sim \alpha^{-1}$, which represents the optimal asymptotic behavior for a model-matched scenario.
These results demonstrate the critical role of the ratio $r$ in determining learning performance. 
Therefore, tuning the ratio $r$ according to the available amount of real data is crucial for achieving optimal performance.  
In practice, it is known that in training GANs, the stability of learning can deteriorate depending on the ratio $r$ of fake to real data.
This theoretical analysis provides insights into the importance of the ratio $r$ and is expected to contribute to improving learning algorithms.

\section{Related Work}\label{sec:related-work}
The replica method is a non-rigorous but powerful heuristic in statistical physics \citep{mezard1987spin, mezard2009information, edwards1975theory}.
It has been proven to be a valuable method for high-dimensional machine-learning problems. 
Previous studies have investigated the relationship between dataset size and generalization error in supervised learning, including single-layer \citep{gardner1988optimal, opper1991generalization, barbier2019optimal, aubin2020generalization} and multi-layer \citep{aubin2018committee} neural networks, as well as kernel methods\citep{dietrich1999statistical, bordelon2020spectrum, gerace2020generalisation}.
In unsupervised learning, the replica method has also been applied to dimensionality reduction techniques such as the principal component analysis \citep{biehl1993statistical, hoyle2004principal, hoyle2007statistical}, and to generative models such as energy-based models \citep{decelle2018thermodynamics, ichikawa2022statistical} and denoising autoencoders \citep{cui2023high}. However, the dataset-size dependence of GANs has not been previously analyzed, which this study aims to address.
Similar to our work, a statistical mechanical formalism for addressing min-max problems has been proposed\citep{varga1998minimax}. 
However, the treatment of the inverse temperature limit differs from our approach, and it has limitations in accurately handling the order of the min and max operations.

\section{Conclusion}\label{sec:conclusion}
This study introduces a statistical mechanical formalism to analyze high-dimensional min-max optimization problems, focusing on the critical order of min and max operations in non-convex scenarios. Our goal was to perform a sensitivity analysis of equilibrium values, providing new insights into their properties and generalization performance.

We applied this approach to a simple min-max game, evaluated the generalization performance of GANs, and derived the optimal ratio of fake to real data for effective learning. This successful application not only validates the approach but also opens the way for extending this formalism to more complex min-max problems and broader applications, suggesting a promising direction for significant advancements in machine learning and optimization.



\subsubsection*{Acknowledgments}
We thank T. Takahashi, K. Okajima, Y. Nagano, and K. Nakaishi for useful discussions and suggestions.
This work was supported by JST Grant Number JPMJPF2221 and JPSJ Grant-in-Aid for Scientific Research Number 23H01095. Additionally, YI was supported by the WINGS-FMSP program at the University of Tokyo.

\bibliography{main}
\bibliographystyle{tmlr}

\appendix
\newpage

\appendix

\section{Derivation of Theorem \ref{theorem:exact-case} proof}
\label{sec:proof-theorem}
In this section, we provide the derivation proof of Theorem \ref{theorem:exact-case}. 
The derivation begins with the calculation of the free energy density without the analytic continuation of $p = -\beta_{\min}/\beta_{\max}$ to $p \in \mab{N}$. 
The free energy density in Eq.~(\ref{eq:non-replica-form}) is connected to the effective Hamiltonian, $\mathcal{H}_{\mathrm{eff}}(\B{x};\B{W})$, which is defined through the relationship: 
\begin{equation*}
    f(\beta_{\min}, \beta_{\max}; \B{W}) = - \frac{1}{\beta_{\min} d_x}\mab{E}_{A} \log \sum_{\B{x}}\exp\left(-\beta_{\min}\mathcal{H}_{\mathrm{eff}}(\B{x};\B{W})\right).
\end{equation*}
The effective Hamiltonian is given by
\begin{align*}
    \mac{H}_{\mathrm{eff}}(\B{x}; \B{W}) &= \frac{1}{\beta_{\max}} \log \sum_{\B{y}} e^{\beta_{\max} V(\B{x}, \B{y}; \B{W})}, \\
    &= \frac{1}{\beta_{\max}} \log \sum_{\B{y}} e^{\beta_{\max} \left(\frac{w_{xx} d_{x}}{2} \big(\frac{\B{x}^{\top} \B{1}_{d_{x}}}{d_{x}} \big)^{2} + \frac{d_{y}w_{yy}}{2} \big(\frac{\B{y}^{\top} \B{1}_{d_{y}}}{d_{y}}  \big)^{2} + w_{xy} \sqrt{d_{x} d_{y}} \big(\frac{\B{x}^{\top} \B{1}_{d_{x}}}{d_{x}} \big) \big(\frac{\B{y}^{\top} \B{1}_{d_{y}}}{d_{y}} \big) \right)} \\
    &\times e^{\beta_{\max} \left( b_{x} d_{x} \big(\frac{\B{x}^{\top}\B{1}_{d_{x}}}{d_{x}} \big) + b_{y} d_{y} \big(\frac{\B{y}^{\top} \B{1}_{d_{y}}}{d_{y}}  \big) \right)}, \\
    &= \frac{1}{\beta_{\max}} \log e^{\beta_{\max} d_{x} \left(\frac{w_{xx}}{2} \big(\frac{\B{x}^{\top} \B{1}_{d_{x}}}{d_{x}} \big)^{2} + b_{x} \big(\frac{\B{x}^{\top}\B{1}_{d_{x}}}{d_{x}} \big)  
    \right)} \\
    &\times \sum_{\B{y}} e^{\beta_{\max} d_{y} \left( \frac{w_{yy}}{2} \big(\frac{\B{y}^{\top} \B{1}_{d_{y}}}{d_{y}}  \big)^{2} + w_{xy} \sqrt{\frac{d_{x}}{d_{y}}} \big(\frac{\B{x}^{\top} \B{1}_{d_{x}}}{d_{x}} \big) \big(\frac{\B{y}^{\top} \B{1}_{d_{y}}}{d_{y}} \big) +  b_{y}  \big(\frac{\B{y}^{\top} \B{1}_{d_{y}}}{d_{y}}  \big)  \right)}, \\
    &= \beta_{\max} d_{x} \left(\frac{w_{xx} d_{x}}{2} \left(\frac{\B{x}^{\top} \B{1}_{d_{x}}}{d_{x}} \right)^{2} +  b_{x} d_{x} \left(\frac{\B{x}^{\top}\B{1}_{d_{x}}}{d_{x}} \right)  
    \right)  \\
    &\times \frac{1}{\beta_{\max}} \log \int d\hat{m}^{y} dm^{y} e^{\beta_{\max} d_{y} \left( \frac{w_{yy}}{2} (m^{y})^{2} + w_{xy} \sqrt{\frac{d_{x}}{d_{y}}} \big(\frac{\B{x}^{\top} \B{1}_{d_{x}}}{d_{x}} \big) m^{y} + b_{y}  m^{y}- \frac{1}{\beta_{\max}} m^{y} \hat{m}^{y} + \frac{1}{\beta_{\max}} \mathrm{Softplus}(\hat{m}^{y}) \right) + o(d_{y})}, 
\end{align*}
where $\mathrm{Softplus}(x) = \log (1+e^{x})$.
To evaluate the integral with respect to $\hat{m}^{y}$ and $m^{y}$, we take the limit as $d_{y}\to \infty$ and apply the saddle point approximation. The effective Hamiltonian can be expressed as follows:
\begin{multline*}
    \mac{H}_{\mathrm{eff}}(\B{x}; \B{W}) = d_{x} \Bigg( \frac{w_{xx}}{2} \left(\frac{\B{x}^{\top} \B{1}_{d_{x}}}{d_{x}}  \right)^{2} + b_{x} \left(\frac{\B{x}^{\top} \B{1}_{d_{x}}}{d_{x}} \right)
    \\
    + \kappa \underset{m^{y}, \hat{m}^{y}}{\mathrm{extr}} \left[\frac{w_{yy}}{2} (m^{y})^{2} + w_{xy} \kappa^{-1/2} \left(\frac{\B{x}^{\top} \B{1}_{d_{x}}}{d_{x}} \right) m^{y} + b_{y}  m^{y}- \frac{1}{\beta_{\max}} m^{y} \hat{m}^{y} + \frac{1}{\beta_{\max}} \mathrm{Softplus}(\hat{m}^{y})  \right] \Bigg).
\end{multline*}
where $\kappa = d_{y}/d_{x}$.
By summing over $\B{x}$, the free energy density can be calculated as follows: 
\begin{align*}
    &f(\beta_{\min}, \beta_{\max}; \B{W}) \\
    &= - \frac{1}{ \beta_{\min}d_{x}} \log \sum_{\B{x}} e^{-\beta_{\min}d_{x} \left( \frac{w_{xx}}{2} \left(\frac{\B{x}^{\top} \B{1}_{d_{x}}}{d_{x}}  \right)^{2} + b_{x} \left(\frac{\B{x}^{\top} \B{1}_{d_{x}}}{d_{x}} \right) \right)}  \\
    &~~~~~~~~~~~\times e^{-\beta_{\min} d_{x} \left(\kappa \underset{m^{y}, \hat{m}^{y}}{\mathrm{extr}} \left[\frac{w_{yy}}{2} (m^{y})^{2} + w_{xy} \kappa^{-1/2} \big(\frac{\B{x}^{\top} \B{1}_{d_{x}}}{d_{x}} \big) m^{y} + b_{y}  m^{y}- \frac{1}{\beta_{\max}} m^{y} \hat{m}^{y} + \frac{1}{\beta_{\max}} \mathrm{Softplus}(\hat{m}^{y})  \right] \right)}, \\
    &= - \frac{1}{\beta_{\min}d_{x}} \log \left(\frac{d_{x}}{2 \pi}  \right) \int dm^{x} d\hat{m}^{x} e^{-\beta_{\min}d_{x}\left(\frac{w_{xx}}{2} (m^{x})^{2} + b_{x} m^{x}  + \frac{1}{\beta_{\min}} m^{x} \hat{m}^{x} - \frac{1}{\beta_{\min}} \mathrm{Softplus}(\hat{m}^{x})  \right)} \\
    &~~~~~~~~~~~\times e^{-\beta_{\min}d_{x} \left(\kappa \underset{m^{y}, \hat{m}^{y}}{\mathrm{extr}} \left[\frac{w_{yy}}{2} (m^{y})^{2} + w_{xy} \kappa^{-1/2} m^{x} m^{y} + b_{y}  m^{y}- \frac{1}{\beta_{\max}} m^{y} \hat{m}^{y} + \frac{1}{\beta_{\max}} \mathrm{Softplus}(\hat{m}^{y})  \right] \right)}, \\
    &= \underset{m^{x}, \hat{m}^{x}, m^{y}, \hat{m}^{y}}{\mathrm{extr}} \Bigg(\frac{w_{xx}}{2} (m^{x})^{2} + b_{x} m^{x} + \frac{1}{\beta_{\min}} m^{x} \hat{m}^{x} - \frac{1}{\beta_{\min}} \mathrm{Softplus}(\hat{m}^{x})  \\
    &~~~~~~~~~~+ \frac{\kappa w_{yy}}{2} (m^{y})^{2} + w_{xy} \kappa^{1/2} m^{x} m^{y} + \kappa b_{y}  m^{y}- \frac{\kappa}{\beta_{\max}} m^{y} \hat{m}^{y} + \frac{\kappa }{\beta_{\max}} \mathrm{Softplus}(\hat{m}^{y})  \Bigg). \\
\end{align*}
The final equality is obtained by applying the saddle point method to evaluate the integral. From the saddle point equations, the following expressions are
\begin{equation*}
    m^{x} = \sigma(\hat{m}^{x}),~~m^{y} = \sigma(\hat{m}^{y})
\end{equation*}
Further transformation of the equation yields the following expression:
\begin{equation}
    \label{eq:complicated-min-max-exact}
    f(\beta_{\min}, \beta_{\max};\B{W}) = \underset{m^{x}, m^{y}}{\mathrm{extr}} \Bigg[ \frac{w_{xx}}{2} (m^{x})^{2} + \frac{\kappa w_{yy}}{2} (m^{y})^{2} + w_{xy} \kappa^{1/2}  m^{x} m^{y} + b_{x} m^{x} + \kappa b_{y} m^{y} 
    - \frac{1}{\beta_{\min}} H(m^{x})+ \frac{\kappa}{\beta_{\max}}H(m^{y})\Bigg],
\end{equation}
where $H(x)= - x \log x - (1-x) \log (1-x)$ represents the binary cross-entropy.

Next, we proceed to evaluate the free energy density under analytic continuation in the replica method, which is expressed as 
\begin{multline*}
    \hat{f}(\beta_{\min}, \beta_{\max}; \B{W}) =- \frac{1}{\beta_{\min} d_{x}} \log \sum_{\B{x}} \sum_{\B{y}_{1}, \ldots, \B{y}_{p}} e^{\beta_{\max} d_{x}  \left(\frac{w_{xx} p}{2} \left(\frac{\B{x}^{\top} \B{1}_{d_{x}}}{d_{x}}   \right)^{2} + \frac{\kappa w_{yy}}{2} \sum_{l} \left(\frac{\B{y}_{l}^{\top} \B{1}_{d_{y}}}{d_{y}}  \right)^{2}   \right)} \\
    \times e^{\beta_{\max} d_{x} \left( w_{xy} \kappa^{1/2} \left(\frac{\B{x}^{\top} \B{1}_{d_{x}}}{d_{x}}  \right) \sum_{l} \left(\frac{\B{y}_{l}^{\top} \B{1}_{d_{y}}}{d_{y}}  \right) + b_{x} p\left(\frac{\B{x}^{\top} \B{1}_{d_{x}}}{d_{x}}  \right) + \kappa b_{y} \sum_{l} \left(\frac{\B{y}_{l}^{\top} \B{1}_{d_{y}}}{d_{y}}  \right)  \right)}.
\end{multline*}
We introduce the order parameter through the Fourier transform representation of the delta function: 
\begin{align*}
    &\hat{f}(\beta_{\min}, \beta_{\max}; \B{W}) \\
    &= - \frac{1}{\beta_{\min} d_{x}} \log  \left( \frac{d_{x} d^{p}_{y}}{(2 \pi)^{p+1}} \right) \int dm^{x} d\hat{m}^{x} \prod_{l} dm_{l}^{y} d\hat{m}_{l}^{y} e^{\beta_{\max} d_{x}  \left(\frac{w_{xx} p}{2} (m^{x})^{2} + \frac{\kappa w_{yy}}{2} \sum_{l} (m^{y}_{l})^{2} + w_{xy} \kappa^{1/2} m^{x} \sum_{l} m_{l}^{y}  \right)} \\
    &\times e^{\beta_{\max} d_{x} \left( b_{x} p m^{x} + \kappa b_{y} \sum_{l}m_{l}^{y} - \frac{1}{\beta_{\max}}\hat{m}^{x} m^{x} - \frac{\kappa}{\beta_{\max}} \sum_{l} \hat{m}^{y}_{l} m^{y}_{l} \right)} \sum_{\B{x}} e^{\hat{m}^{x} \sum_{i} x_{i}} \sum_{\B{y}_{1}, \ldots, \B{y}_{p}} e^{\sum_{l} \hat{m}_{l}^{y} \sum_{j} y_{j}}, \\
    &= - \frac{1}{\beta_{\min} d_{x}} \log  \left( \frac{d_{x} d^{p}_{y}}{(2 \pi)^{p+1}} \right) \int dm^{x} d\hat{m}^{x} \prod_{l} dm_{l}^{y} d\hat{m}_{l}^{y} e^{\beta_{\max} d_{x}  \left(\frac{w_{xx} p}{2} (m^{x})^{2} + \frac{\kappa w_{yy}}{2} \sum_{l} (m^{y}_{l})^{2} + w_{xy} \kappa^{1/2} m^{x} \sum_{l} m_{l}^{y}  \right)} \\
    &\times e^{\beta_{\max} d_{x} \left( b_{x} p m^{x} + \kappa b_{y} \sum_{l}m_{l}^{y} - \frac{1}{\beta_{\max}}\hat{m}^{x} m^{x} - \frac{\kappa}{\beta_{\max}} \sum_{l} \hat{m}^{y}_{l} m^{y}_{l} + \frac{1}{\beta_{\max}} \mathrm{Softplus}(\hat{m}^{x}) + \frac{\kappa}{\beta_{\max}} \sum_{l} \mathrm{Softplus}(\hat{m}_{l}^{y})\right)}. 
\end{align*}
Under the assumption of replica symmetry, where $\forall l \in [p],~~\hat{m}_{l}^{y}=m^{y}, m_{l}^{y}=m^{y}$, we can further reformulate the expression as follows:
\begin{align*}
    &\hat{f}(\beta_{\min}, \beta_{\max}; \B{W}) \\
    &= - \frac{1}{\beta_{\min} d_{x}} \log  \int dm^{x} d\hat{m}^{x} dm^{y} d\hat{m}^{y} e^{-\beta_{\min}  d_{x}  \left(\frac{w_{xx}}{2} (m^{x})^{2} + \frac{\kappa w_{yy}}{2} (m^{y})^{2} + w_{xy} \kappa^{1/2}  m^{x} m^{y}  \right)} \\
    &~~~~~~~~~~\times e^{-\beta_{\min}  d_{x} \left( b_{x} m^{x} + \kappa b_{y} m^{y} - \frac{1}{\beta_{\max}p}\hat{m}^{x} m^{x} - \frac{\kappa }{\beta_{\max}} \hat{m}^{y} m^{y} + \frac{1}{\beta_{\max}p} \mathrm{Softplus}(\hat{m}^{x}) + \frac{\kappa }{\beta_{\max}}  \mathrm{Softplus}(\hat{m}^{y}) + o(d^{x}) + o(d^{y}) \right)}, \\ 
    &= \underset{m^{x}, m^{y}, \hat{m}^{x}, \hat{m}^{y}}{\mathrm{extr}}\Bigg[ \frac{w_{xx}}{2} (m^{x})^{2} + \frac{\kappa w_{yy}}{2} (m^{y})^{2} + w_{xy} \kappa^{1/2}  m^{x} m^{y} + b_{x} m^{x} + \kappa b_{y} m^{y} \\
    &~~~~~~~~~- \frac{1}{\beta_{\max}p}\hat{m}^{x} m^{x} - \frac{\kappa }{\beta_{\max}} \hat{m}^{y} m^{y} + \frac{1}{\beta_{\max}p} \mathrm{Softplus}(\hat{m}^{x}) + \frac{\kappa }{\beta_{\max}}  \mathrm{Softplus}(\hat{m}^{y}) \Bigg].
\end{align*}
The final equality is derived by handling the integral using the saddle point method. 
Consequently, the following saddle point equation is obtained:
\begin{align*}
    m^{x} &= \sigma(\hat{m}^{x}),~~m^{y} = \sigma(\hat{m}^{y}).
\end{align*}
Substituting these results, the following expression for the free energy density is derived as 
\begin{multline}
    \hat{f}(\beta_{\min}, \beta_{\max}; \B{W}) = \underset{m^{x}, m^{y}}{\mathrm{extr}}\Bigg[ \frac{w_{xx}}{2} (m^{x})^{2} + \frac{\kappa w_{yy}}{2} (m^{y})^{2} + w_{xy} \kappa^{1/2}  m^{x} m^{y} + b_{x} m^{x} + \kappa b_{y} m^{y}
    - \frac{1}{\beta_{\min}} H(m^{x}) +\frac{\kappa}{\beta_{\max}} H(m^{y})\Bigg].
\end{multline}
This result coincides with the exact free energy density $f(\beta_{\min}, \beta_{\max}; \B{W})$, derived without the need for analytic continuation.

\end{document}